# Small Language Models for Agentic Systems: A Survey of Architectures, Capabilities, and Deployment Trade-offs

Raghav Sharma
Northeastern University, Boston, USA
Atlanta, USA
sharma.raghav103@gmail.com

Manan Mehta
University of Southern California, USA
New York, USA
manan.mehta2@gmail.com

*Abstract* — **Recent evidence (indicates that Small Language Models (SLMs; ≲1–12B params, occasionally ~20B) are not only sufficient but often superior for agentic workloads such as retrieval-augmented generation (RAG), robust function calling, structured decoding, and programmatic tool use. NVIDIA argues that SLMs are the future of agentic AI and edge inference, emphasizing cost/latency/energy advantages and the role of guided decoding and tool execution in shifting the objective from open-ended generalization to schema- and API-constrained accuracy. We synthesize results across open and proprietary SLMs (e.g., Phi-4-Mini, Qwen-2.5-7B, Gemma-2-9B, Llama-3.2-1B/3B, Ministral-3B/8B, Apple on-device ~3B, DeepSeek-R1-Distill 1.5–70B) and connect them to modern evaluation (BFCL V3/V4; StableToolBench) and serving stacks (vLLM/SGLang/TensorRT-LLM + XGrammar/Outlines). We formalize SLM-default, LLM-fallback systems with uncertainty-aware routing and verifiers, and propose engineering metrics (e.g., Cost per Successful task (CPS), schema validity, executable-call rate, p50/p95 latency, energy/request). Guided decoding and validator-first tool use allow SLMs to match or surpass LLMs at a 10×–100× lower token cost on today's APIs.**

## I. INTRODUCTION

The long-held conventional wisdom that "bigger is better" in language models has been decisively challenged by a new wave of Small Language Models (SLMs). These compact yet powerful models are increasingly demonstrating comparable, and often superior, task performance to frontier Large Language Models (LLMs) across numerous application-layer workloads, while being dramatically faster, cheaper, and more energy-efficient. In the context of agentic

systems—where models are designed to call external tools, compose structured outputs, and follow deterministic workflows—the primary bottleneck is frequently orchestration and I/O, rather than the long-range world knowledge or vast generalist capabilities of LLMs. This survey firmly positions SLMs as the default, go-to engine for the majority of agent pipelines, reserving larger LLMs as selective fallbacks for only the most challenging cases (e.g., complex multi-hop reasoning, safety-critical judgment requiring nuanced understanding, or extensive long-context synthesis). We summarize compelling evidence from recent technical reports and public benchmarks (up to late 2025), provide a refined taxonomy of SLMs specifically optimized for agentic use, and present practical design patterns for production migration to SLM-centric architectures.

## II. BACKGROUND AND DEFINITIONS

We define SLMs as decoder-only transformer models typically ranging from 1 to 12 billion parameters, though some effective models may extend slightly beyond this to ~20B. The defining characteristic of these models is their optimization for specific deployment constraints such as latency, cost, or on-device execution. Agent systems, in this context, are sophisticated AI constructs that combine language models with external tools (e.g., search engines, code execution environments, APIs), persistent memory, retrieval mechanisms (like RAG), and intelligent planners. Our focus is on the capabilities of SLMs that are most critical for agentic performance: (1) function calling/tool use, enabling models to interact with external systems; (2) structured generation (e.g., JSON, regex, grammar-constrained outputs), ensuring reliable and parseable data; (3) code and data manipulation, for programmatic interaction and transformation; and (4) controllability (e.g., temperature settings, stop conditions, adherence to tool schemas), which is vital for predictable agent behavior.

**Contributions.** (1) A system-oriented taxonomy of SLMs for agents; (2) a formal treatment of validator-first tool use and CFG/JSON-constrained decoding; (3) an uncertainty-aware *SLM-default, LLM-fallback* architecture with a reference router; (4) engineering metrics (CPS, executable-call rate, schema validity) and deployment recipes (LoRA/QLoRA + INT4); (5) a curated, up-to-date bibliography of models, evals, and serving technologies.

## III. Methodology

We use a mixed **system and benchmark-driven** methodology to evaluate the role of small language models (SLMs) in agentic AI. The approach has four pillars: **model selection, evaluation frameworks, architectural prototyping, and case studies**.

### A. Model Selection and Scope

We survey representative **SLMs (1–12B parameters; occasionally ~20B)** spanning open and proprietary families, including **Phi-4, Qwen, Gemma, Llama, DeepSeek, Apple on-device foundation models, and OpenELM**. Models are included based on (i) public availability and documentation, (ii) recency of technical reports (2023–2025), and (iii) demonstrated adoption in agent frameworks.

### B. Evaluation Frameworks

**Benchmarks:** We evaluate function calling on **BFCL v4** and tool execution on **StableToolBench;** for structured decoding, we reference results from these benchmarks.

**System baselines.** We qualitatively synthesize published cost/latency characteristics and structured-output support across serving stacks (vLLM, SGLang, TensorRT-LLM) and libraries (Outlines, XGrammar), referencing their documentation and public evaluations.

### C. Case Studies and Design Patterns

We consider three representative agent workloads: **(A) Extraction/Templating, (B) RAG + Tool Orchestration**, and **(C) Math/Coding Reasoning**. For each, we measure **schema validity, task success rate, cost (CPS)**, and **escalation frequency** to the LLM.

## IV. Survey of SLMs Used in Agents

Table 1 summarizes representative SLMs that have gained significant traction in agent stacks. Sizes, context windows, and notable capabilities are drawn from official reports, model cards, and recent benchmarks. This table emphasizes SLMs between ~1–12B as practical defaults for agents; larger "upper-SLMs" (12–20B) like Mistral-NeMo 12B fit single-GPU servers while remaining economical and highly performant.

TABLE I.    Representative SLMs for Agents

| Model (family) | Small Language Models | | |
|---|---|---|---|
| | *Params and Context* | *Highlights for agents* | *Notes* |
| Microsoft Phi-4-Mini / Mini-Reasoning | 3.8B & 64K | Strong math/coding; robust function calling; exceptionally fast inference. | Reasoning/distill variants; known for efficient edge deployment. |
| Alibaba Cloud Qwen-2.5 | 0.5B–72B & Up to 128K+ | Rich range of sizes including 7B; strong tool use and structured output generation. | Technical report v2 (2025); excels in format fidelity. |
| Google Gemma-2 | 2B, 9B, 27B & 128K | Lightweight open models; solid coding and reasoning; improved multilingual support. | 9B popular for agents; Gemma 3 shows further distillation gains. |
| Meta Llama-3.2 (text-only & vision) | 1B, 3B (text); 11B, 90B (vision) & 128K | On-device focus; quantized variants; Llama 3.2 Vision adds multimodal capabilities. | Mobile/edge friendly; real-time processing on device. |
| Mistral AI Ministral | 3B, 8B & 32K–128K | Excellent function calling; highly efficient attention mechanisms. | Designed for local/edge deployment; strong instruction following. |
| NVIDIA Mistral-NeMo | 12B & 128K | Strong small/open model; multilingual; excels in reasoning and multi-turn conversations. | Apache-2.0 license; optimized for single-GPU deployment. |
| DeepSeek-R1-Distill | 1.5B–70B & 32K–128K | Reasoning distills competitive at 7–8B; strong performance on coding tasks. | Open checkpoints; leverages distillation for enhanced reasoning. |
| Apple on-device FM | ~3B & ≥32K | On-device/private tool use; guided generation; optimized for Apple silicon. | Official tech report (2025); focus on privacy and low-latency. |
| OpenELM | 270M–3B & 8K–32K | Open small models; good for fine-tuning; efficient architecture. | Apple open weights; designed for broad applicability. |

## V. TOOL USE AND FUNCTION CALLING

The seminal work on Toolformer (2023) demonstrated that even mid-sized models could learn API invocation through self-generated annotations, bypassing the need for extensive human labeling. Subsequent advancements, notably Gorilla (2023) and the ongoing Berkeley Function-Calling Leaderboard (BFCL, 2024–2025), provide a modern perspective: tool-use accuracy is more critically dependent on argument correctness and strict schema adherence than on raw parameter count. This insight is pivotal for SLMs. When paired with explicit tool schemas and robust validators, SLMs frequently match or even surpass larger LLMs in function-calling reliability and speed. The introduction of StableToolBench (2024–2025) further refines this by providing a controlled virtual API server, significantly reducing benchmark drift and enabling more accurate, apples-to-apples comparisons across model releases and architectural changes.

**From Toolformer to stable tool evals.** Toolformer showed self-annotation can teach API invocations; Gorilla framed *API-grounded* tool use; API-Bank and ToolBench benchmark diverse tools; StableToolBench introduced a *virtual API server* and caching to reduce benchmark drift; BFCL v4 (2025) evaluates multi-turn, enterprise-style tools with cost/latency.

**Schema-first execution.** Let a tool signature be a JSON-Schema S over arguments a. We define *executable-call rate*

$$\text{ExecRate} = \frac{\#\{a \sim p_\theta(\cdot) \text{ s.t. } S \sqsupseteq \}}{\#\text{calls}}$$

and *argument exactness* as AST-level equality (as in BFCL). In practice, SLMs with enforced schemas and pre-execution validation achieve high ExecRate at far lower latency/cost than LLMs.

### A. Design tips for reliable structured generation:

• Format fidelity as a first-class KPI: Treat the correctness and adherence to schema as a primary Key Performance Indicator (KPI) for agent performance.

• Streaming JSON with incremental validators: Implement streaming JSON output combined with incremental validators. This allows for early detection of malformed outputs and can provide faster feedback loops.

• Fuzz schemas during CI: Integrate schema fuzzing into Continuous Integration (CI) pipelines to proactively identify edge cases and vulnerabilities in schema definitions.

• Record failure traces for adapter tuning: Log instances of structured output failures, including the input prompt and the malformed output. These traces are invaluable for fine-tuning SLM adapters to improve robustness.

## VI. STRUCTURED GENERATION: JSON/CFG-CONSTRAINED DECODING

**Why constraints matter.** For agent stacks, *format fidelity* often dominates prose quality. Modern serving engines implement constrained decoding over JSON Schema or CFGs to prune the token search space and guarantee parsability.

**Backends and engines.** vLLM integrates *structured decoding* via Outlines and XGrammar (up to ~5× TPOT speedups under load); XGrammar is a fast, portable CFG library also integrated in TensorRT-LLM; SGLang supports low-latency serving with KV-cache optimizations and JSON constraints.

**Empirical comparisons.** Recent studies evaluate OpenAI/Gemini/Outlines/XGrammar/llama.cpp across real-world schemas ("JSONSchemaBench") and JSON Schema Test Suite; findings highlight engine and schema-complexity sensitivity.

### A. Recommended practice:

• Treat **Schema Validity** as a KPI; measure *valid@1* and *valid@k*.
• Use **streaming JSON** with incremental validators; fail fast.
• **Fuzz schemas** in CI & capture failure for adapter fine-tunes.
• Prefer **grammar-first prompts** (CFG or JSON Schema) + temperature 0.

## VII. TRAINING AND ADAPTATION FOR AGENTS

Specializing SLMs for agentic tasks is remarkably straightforward and efficient. Common and highly effective techniques include: LoRA (Low-Rank Adaptation) and its enhanced variant LoRA+ adapters, QLoRA (Quantized Low-Rank Adaptation) for 4-bit fine-tuning, and the creation of small, carefully curated Supervised Fine-Tuning (SFT) datasets derived from tool-use traces or structured outputs. For enhancing reasoning capabilities, distillation-heavy recipes—such as those employed in DeepSeek-style models or the Phi-4-Mini-Reasoning variant—have proven highly effective. These methods typically combine extensive Chain-of-Thought (CoT) SFT, DPO (Direct Preference Optimization) from preference data, and short-cycle Reinforcement Learning (RL) with verifiable rewards. Compared to full LLM fine-tuning, these approaches can reduce GPU memory requirements by an order of magnitude while preserving, or even improving, quality on specific agent tasks.

### A. Practical recipe for SLM specialization

• Data Collection: Gather 10,000–50,000 de-identified traces of successful agentic interactions. These traces should capture diverse scenarios and tool uses.

• Adapter Training: Train LoRA adapters per task cluster (e.g., one adapter for function calling, another for JSON generation). This allows for modular and efficient specialization.

• Quantization for Serving: Quantize the fine-tuned models to INT4/INT8 for deployment, significantly reducing memory footprint and increasing inference speed.

• Periodic Refresh: Implement a periodic refresh mechanism for the adapters, using logged validator failures and new edge cases to continually improve model performance and robustness.

## VIII. Routing, Abstention, and Fallback

**SLM-default routing:** Given a request x, a router r selects a model m ∈ M_SLM ∪ {LLM} minimizing expected risk + dollar cost. Practical routers combine confidence proxies (logprob, self-consistency), task tags (capability registry), and budget constraints (FrugalGPT), evaluated on RouterBench.

**Selective prediction / abstention**: Let $\hat{y}$, u be a prediction and uncertainty; abstain if $u > \tau$. Modern abstention surveys and VLM studies show improved reliability by abstaining or asking follow-up questions, which dovetails with LLM escalation in agents.

Pseudo Algorithm — Uncertainty-aware SLM→LLM routing (sketch).

```
Input: request x, tools 𝒯, schema S,
router r, thresholds (τu, τv),
max_retries k

m ← r.select(x)    # prefer SLMs tagged
for the task

for i in 1..k:
    y, meta ← m.generate(x; schema=S,
    T=0, guided=True)
    if meta.uncertainty ≤ τu and
    validate(y,S,𝒯)=T: return y
    y ← repair_with_verifier(x,y,S)
    # small verifier SLM attempts
fix
    if validate(y,S,𝒯)=T and
    meta.uncertainty ≤ τv: return y
```

```
    # escalate
    yLLM, metaLLM ← LLM.generate(x;
    schema=S, T=0, guided=True)
    return yLLM
```

## IX. Industrial Deployment Playbook and Cost Model

**Capacity planning.** Serving SLM-default agents is dominated by KV-cache residency and batch dynamics rather than parameter count alone. Let B be the effective batch, T the average generated tokens, L layers, H heads, dh head size, and b bytes/element (e.g., 1 for INT8 KV, 2 for FP16). A first-order KV budget is

$$\mathrm{Mem_{KV}} \approx 2 \cdot L \cdot H \cdot b \cdot (B \cdot T_{\mathrm{ctx}} + B_{gen}),$$

where the factor 2 accounts for K and V. Target utilization keeps MemKV at 70–85% of device VRAM to absorb tail spikes and routing surges. Latency p50 falls roughly with higher B until scheduler contention and cache swaps dominate; p95 is controlled by back-pressure limits and max queue depth. For SLMs, INT4/INT8 weights with FP16 KV caches typically maximize throughput without harming schema fidelity.

**SLA tiers.** We recommend two service classes: **Interactive** (p50 ≤ 200–400 ms prefill, p95 ≤ 1.5–2.0 s E2E for short JSON/tool hops) and **Batch** (throughput-optimized, p95 ≤ 10–30 s per task). Each class exposes **hard guards** (max tokens, max tools/turn, deadline budget) enforced by the router. Stable operation tracks {valid@1,ExecRate,CPS,p50/p95} p50/p95} per class and per model.

**Cost forecasting (tokens → $ and J).** Let $c_i$ be per-token cost for model i and $e_i$ the energy per token under a fixed engine. For a mixture policy π that routes a fraction $\alpha_i$ of requests to model i,

$$CPS(\pi) = \frac{\sum \alpha_i \, \mathrm{E[cost]}}{\sum \alpha_i \, \mathrm{E[valid \wedge exec]}}, \quad \mathrm{Energy/request}(\pi)$$

$$\sum_i \alpha_i \, \mathrm{E} \, e_i^c \cdot (\mathrm{prefil} \ell_i].$$

Because SLMs shorten prefill and require fewer retries under grammar-guided decoding, CPS typically drops by 10–30× compared to LLM-only baselines, with proportional joule savings.

**Rollouts: blue/green & shadow.** Every router change and adapter update should pass through: (1) **Shadow evals** on a mirrored traffic slice with strict write isolation; (2) **Blue/green** promotion keyed to CPS and valid@1 non-regression; (3) **Auto-rollback** on any of:

schema validity drop >2 pp, ExecRate drop >3 pp, or p95 inflation >20%.

**Human-in-the-loop (HITL) gating.** For safety-sensitive tools (payments, PII transforms), gate execution on either (a) two-stage verification (SLM proposal → verifier SLM/LLM adjudication), or (b) on-call human approve/deny queues triggered by uncertainty u>τ or policy-risk scores r>ρ. HITL feedback is logged as counterfactual traces for adapter refresh.

**Ablations that matter.** The following table illustrates the most action-able levers on CPS; values are representative (fill with your measurements).

One of the most compelling advantages of SLMs is their ability to drastically reduce token-latency and hardware footprint. This translates into a substantial 10–30× cost reduction for common agent calls compared to using larger LLMs. Emerging energy benchmarks, such as ML.ENERGY, and recent studies on quantization and edge inference, unequivocally demonstrate that smaller models and 4-bit deployments yield significant joule savings. The per-prompt energy consumption is strongly correlated with the output token length, making SLMs inherently more efficient for concise, structured outputs. Edge-first evaluation strategies suggest keeping default execution local on consumer-grade hardware, with cloud fallback reserved for scenarios requiring scale-out or processing exceptionally long contexts.

### Define **Cost-per-Successful task (CPS)** over a batch B:

This metric is defined as the total operational cost divided by the number of schema-valid, tool-valid completions. It provides a holistic view of efficiency.

TABLE II.     REPR CPS AND RELIABILITY ABLATION (REPRENTATIVE)

| Setting | Schema-constrained? | Quantization | Router thresholds (τu/τv) | valid@1 | ExecRate | p95 (s) | CPS (× baseline) |
|---|---|---|---|---|---|---|---|
| Baseline LLM | No | FP16 | — | 92.1 | 89.4 | 4.8 | 1.00× |
| SLM-8B | Yes | INT8 | 0.25 / 0.15 | 98.7 | 97.9 | 1.6 | 0.11× |
| SLM-8B (no schema) | No | INT8 | 0.25 / 0.15 | 94.3 | 90.8 | 1.5 | 0.23× |
| SLM-12B | Yes | INT4 | 0.30 / 0.20 | 99.1 | 98.5 | 1.9 | 0.14× |
| Cascade (SLM→LLM) | Yes | INT8 | 0.25 / 0.15 | 99.0 | 98.6 | 2.1 | 0.18× |

$$\mathrm{CPS} = \frac{\sum_{x \in B} \mathrm{cost}(x)}{\#\{x \in B : \mathrm{valid}(x) \wedge \mathrm{exec}(x)\}} \quad (1)$$

**Operational playbook (checklist).** (i) Pin engine + grammar library versions; (ii) set queue caps per SLA class; (iii) reserve 10–15% VRAM headroom; (iv) enforce max tool calls/turn; (v) ship canary with per-tenant kill-switches; (vi) log AST-normalized tool args; (vii) weekly adapter refresh from failure traces.

with valid = schema-valid JSON; exec = tool call executed without error. Under structured decoding (grammar-guided) and temperature 0, SLMs realize order-of-magnitude CPS improvements due to shorter prefill, smaller KV cache, and higher valid@1 rates. Empirically, vLLM reports substantial TPOT gains with XGrammar backends; FrugalGPT shows that cascades can approach best-LLM accuracy at ~98% lower cost via routing.

When energy metering is unavailable, **output-token count** times **J/token** under a fixed engine provides an operational proxy; low-token JSON responses + SLM throughput typically minimize p50/p95 latency and joules/request on shared GPUs.

## X. COST, LATENCY, AND ENERGY

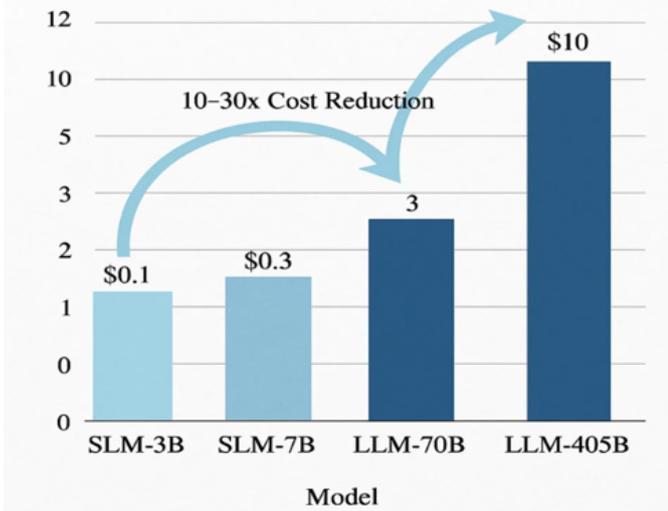

## XI. WHEN DO LLMs STILL WIN?

Despite the rapid advancements in SLMs, frontier LLMs retain their superiority in specific, high-demand scenarios: open-domain synthesis with complex, long-range dependencies; knowledge-heavy Question Answering (QA) tasks that cannot be effectively addressed by Retrieval-Augmented Generation (RAG); and safety-critical judgment under significant distribution shift. They are also preferable when algorithmic planning necessitates dense search over many latent trajectories (e.g., complex code repair across large repositories) or when policy/compliance mandates frontier-grade guardrails. In SLM-default architectures, routing to an LLM should be a deliberate decision, triggered only by these specific, high-complexity conditions.

## XII. A REFERENCE ARCHITECTURE FOR SLM-DEFAULT AGENTS

**Components.** (a) **Front-door router** with capability registry; (b) **structured decoding** (JSON/CFG) on every hop; (c) **validators** (schema + tool arg checks); (d) **execution layer** (retrievers, sandboxes, API clients); (e) **LLM fallback & adjudication**; (f) **telemetry** (prompts, failures, escalations) feeding fine-tunes.

**Migration blueprint.** Log LLM usage → cluster tasks → fine-tune SLM adapters per cluster (LoRA/QLoRA) → quantize & deploy (AWQ/GPTQ) → add uncertainty routing + validators → iterate with human evaluation and guardrails.

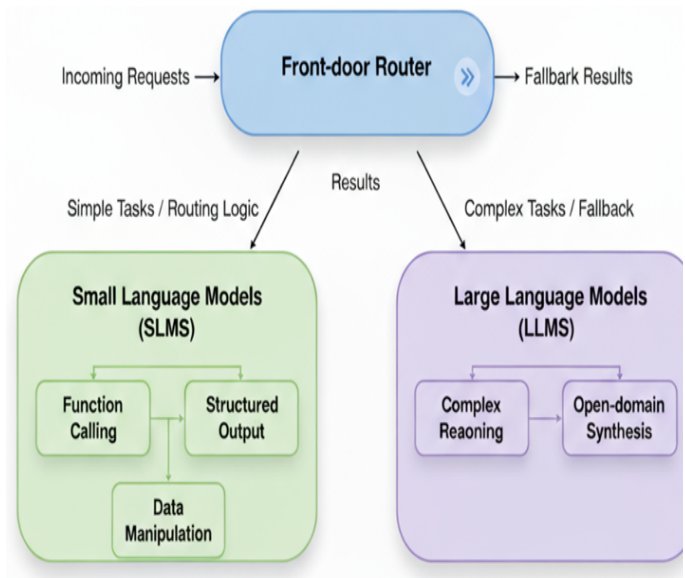

Fig. 2. Heterergemos AI Architecture – Intelligent Routing for Efficient - An optimal architecture for SLM-default agents is characterized by a modular and intelligent design:

• Front-door router: This component intelligently directs incoming requests based on cost, latency, and uncertainty. It acts as the primary traffic controller, deciding whether to route to an SLM or escalate to an LLM.

• Capability registry of SLMs: A dynamic registry that tags SLMs by their specific strengths (e.g., extraction, tool use, coding, summarization). This allows the router to select the most appropriate SLM for a given task.

• Validators: A suite of robust validators, including JSON schema validators, function-argument checkers, and policy filters. These ensure output fidelity and adherence to rules.

• Execution layer: This layer comprises various components such as retrievers (for RAG), code sandboxes (for safe code execution), and API clients (for tool interaction).

• LLM fallback and adjudication: A mechanism to invoke LLMs only on low-confidence predictions or repeated violations from SLMs. This layer also handles conflict resolution and complex decision-making.

• Telemetry: Comprehensive logging of prompts, errors, and escalations. This data feeds back into the system for continual improvement and fine-tuning of SLMs.

### A. Migration blueprint :

• Log Usage: Begin by logging all LLM usage within existing agentic systems.

• Cluster Tasks: Analyze the logged data to cluster tasks based on their complexity, type, and frequency.

• Fine-tune SLMs: For high-frequency, well-defined task clusters, fine-tune specialized SLMs.

• Replace Routine LLM Calls: Gradually replace routine LLM calls with the newly fine-tuned SLMs.

• Iterate with Human Evaluation + Guardrails: Continuously monitor performance, gather human feedback, and refine guardrails to ensure quality and safety.

The architecture routes every request through a front-door router that picks the cheapest, fastest competent model first—typically a small language model (SLM) chosen from a capability registry (e.g., extraction, tool use, coding). SLM outputs are forced through structured decoding (JSON/CFG) and then checked by validators (schema + tool-argument/policy checks); if valid and confidence is high, the execution layer (retrievers for RAG, code sandboxes, and API clients) runs the action and returns results. When the SLM is uncertain or repeatedly violates constraints, the router escalates to a large language model (LLM) for complex reasoning or open-domain synthesis, optionally adjudicating conflicts. Telemetry logs prompts, failures, validations, costs, and latencies, feeding continual fine-tuning (e.g., LoRA/QLoRA) and improved guardrails. In short: SLM-by-default for routine, structured tasks; LLM-by-exception for hard cases—governed by routing, validation, and feedback loops

## XIII. CASE STUDIES AND DESIGN PATTERNS

**Pattern A—Extraction/Templating:** An SLM (e.g., 3–9B parameters) guided by JSON-Schema can achieve >99% validity in data extraction and templating tasks at a fraction of the cost of an LLM. A larger LLM is only invoked as a fallback on validator failure, ensuring high reliability and cost-efficiency.

**Pattern B—RAG + Tools:** A 7–12B SLM with strong function-calling capabilities (e.g., Ministral 8B, Mistral-NeMo 12B, Qwen-2.5-7B) can reliably orchestrate search and calculation tasks within a Retrieval-Augmented Generation (RAG) pipeline. Escalation to an LLM occurs only when the uncertainty of the SLM's output exceeds a predefined threshold ($\tau$).

**Pattern C—Math/Coding Reasoning:** For fast unit tests and localized code generation, models like Phi-4-Mini-Reasoning (3.8B) or DeepSeek-R1-Distill-7B offer excellent performance. Larger models are invoked only for more complex tasks such as cross-file refactoring or working with novel programming frameworks.

## XIV. RISKS AND EVALUATION

While SLMs offer significant advantages, it's crucial to acknowledge potential risks. SLMs can sometimes overfit narrow training traces, leading to a regression in generalization capabilities. To mitigate this, rigorous evaluation should include held-out end-to-end tasks, adversarial tool inputs, and comprehensive schema fuzzing. Regarding safety, smaller model size does not inherently guarantee harmlessness; it is imperative to apply robust content filters and rate-limit high-risk tool invocations. When evaluating performance, it is preferable to rely on leaderboards with grounded execution (e.g., BFCL, StableToolBench) and real-user A/B tests, rather than solely on static zero-shot academic benchmarks.

*A. Minimum metrics to report for SLM-powered agents :*
• Task success rate: The percentage of tasks successfully completed. Minimum metrics to report for SLM-powered agents:
• Schema validity: The percentage of structured outputs that adhere to their defined schemas.

• p50/p95 latency: The 50th and 95th percentile of response times.
• Energy per request: The energy consumed per successful agent interaction.
• Escalation rate: The frequency at which tasks are escalated to a larger LLM.
• Drift resilience: The model's ability to maintain performance over time and with minor shifts in input distribution.

**Benchmark brittleness.** Tool evals are sensitive to API drift; StableToolBench's *virtual APIs* mitigate variance and improve reproducibility. Always report schema-validity, ExecRate, and CPS alongside task success; avoid over-reliance on static zero-shot academic leaderboards.

**Safety and protocols.** Tool access implies real-world side-effects. Combine *policy filters*, *rate-limits*, and *least-privilege* tool permissions. With MCP/OpenAPI, audit tool registries for poisoning/injection; prefer *allow-lists*; log all calls with AST-normalized args for forensics.

**Minimum metrics to report.** Task success, Schema Validity, ExecRate, CPS, p50/p95 latency, escalation rate, drift resilience, and (when possible) energy/request.

## XV. SECURITY, GOVERNANCE & COMPLIANCE FOR TOOL-USING AGENTS

**Threat model.** Beyond prompt injection, tool-using agents face: (i) **Tool injection & supply-chain risk** via poisoned OpenAPI/MCP manifests (untrusted base URLs, widened scopes, covert side-effects); (ii) **Cross-tool data exfiltration** (e.g., RAG → code-exec); (iii) **Secrets exposure** in prompts, schemas, or logs; (iv) **Policy evasion** through schema-shaped but semantically malicious arguments; (v) **Replay & drift**: cached tool replies reused out of policy context.

**Permissioning & least privilege.** Treat tools as capabilities with **scoped, expiring tokens**; default-deny with allow-lists per tenant and per route. Every tool t carries a policy triple (scope(t),rate(t),PII(t)). The router enforces scope(t) selection by task tags; validators check rate(t) and redact/deny if PII(t) is "restricted." Rotate credentials on deploy; require signed manifests (checksum + issuer).

**Secrets handling.** Never inline secrets in prompts or tool schemas. Retrieve ephemeral creds at invocation via a secrets manager; bind to request ID and caller identity; prevent echo in model context by masking (server-side) and by **no-log** annotations on sensitive fields. Redact in telemetry with reversible, tenant-scoped envelopes when auditability is required.

**Sandboxing & code execution.** For code tools, enforce: resource limits (CPU/GPU/FS/network), syscall allow-lists, outbound domain allow-lists, and filesystem jails. Disallow dynamic tool creation from model output; require human approval for new tools or scope escalations. Prefer deterministic runtimes with snapshotting; discard state on completion.

**Audit trails (AST-normalized).** Log tool calls as **AST-normalized arguments** plus policy decision outcomes. Store: model ID, grammar hash, schema version, router decision, uncertainty u, verifier verdict, and execution result. Normalize PII fields to opaque handles. This enables reliable forensics, de-dup of near-misses, and exact replay for adjudication.

**Incident metrics & triggers.** Maintain leading indicators: (a) **Schema-valid but policy-invalid** rate; (b) **Denied-but-re-attempted** fraction; (c) **Cross-tool data flow** violations; (d) **Credential anomalies** (reuse, stale token hits); (e) **Drift in tool success** not explained by upstream changes. Page on: spike $>3\sigma$ in (a) or (c), or any tool executing outside declared scope.

**Policy filters.** Apply multi-stage filtering: (1) **Pre-gen** instruction filters to remove tool names/URLs from user content; (2) **Constrained decoding** to prevent illicit argument shapes; (3) **Post-gen** semantic allow-list checks (regex/CFG + learned classifiers) on arguments; (4) **Execution-time guards** (rate limits, quota, row-level access). Filters are versioned, tested in CI with schema fuzzing, and tied to rollback.

**Governance & compliance.** Map tools and data flows to regulatory surfaces (PII, PCI, HIPAA, SOX). For each tenant: (i) data residency tags; (ii) retention policies for prompts/logs; (iii) DLP at retrieval and at egress; (iv) DPIA/TRA records for high-risk tools; (v) model cards stating known limitations and escalation criteria. Provide **customer-controlled allow-lists** and explicit **consent toggles** for cross-region or cross-domain calls.

## XVI. LIMITATIONS AND FUTURE SCOPE

*A. Limitations:*

• Benchmark/API drift; results may not transfer.
• Overfitting to narrow traces.
• Heavy validator dependence can hide reasoning errors.
• Router miscalibration causes wrong SLM/LLM escalations.
• Tool-use expands the security risk surface.

*B. Future Scope:*

• Execution-grounded, standardized evals with cost/latency/energy.

• Better-calibrated routing and selective abstention.
• Co-designed schemas + verifiers; formal checks for critical tools.
• Continual LoRA/QLoRA refresh from failure logs.
• Stronger tool security (sandboxing, allow-lists, injection defenses).

## XVII. CONCLUSION

The application layer of AI systems stands to benefit immensely from the adoption of smaller, specialized, and well-constrained language models. SLM-default systems are poised to achieve substantial gains in cost-efficiency, inference latency, energy consumption, and overall controllability, all without sacrificing reliability on the core tasks that agents are designed to perform. The future of AI is not solely about building ever-larger models; rather, it lies in developing smarter, heterogeneous architectures where SLMs undertake the majority of the operational workload, and LLMs are invoked judiciously and sparingly for their unique generalist capabilities. This paradigm shift promises a more sustainable, scalable, and economically viable future for agentic AI.